# High Performance Offline Handwritten Chinese Character Recognition Using GoogLeNet and Directional Feature Maps

Zhuoyao Zhong, Lianwen Jin+, Zecheng Xie
School of Electronic and Information Engineering, South China University of Technology
Guangzhou, China
z.zhuoyao@mail.scut.edu.cn, +Lianwen.jin@gmail.com

***Abstract*—Just like its great success in solving many computer vision problems, the convolutional neural networks (CNN) provided new end-to-end approach to handwritten Chinese character recognition (HCCR) with very promising results in recent years. However, previous CNNs so far proposed for HCCR were neither deep enough nor slim enough. We show in this paper that, a deeper architecture can benefit HCCR a lot to achieve higher performance, meanwhile can be designed with less parameters. We also show that the traditional feature extraction methods, such as Gabor or gradient feature maps, are still useful for enhancing the performance of CNN. We design a streamlined version of GoogLeNet [13], which was original proposed for image classification in recent years with very deep architecture, for HCCR (denoted as HCCR-GoogLeNet). The HCCR-GoogLeNet we used is 19 layers deep but involves with only 7.26 million parameters. Experiments were conducted using the ICDAR 2013 offline HCCR competition dataset. It has been shown that with the proper incorporation with traditional directional feature maps, the proposed single and ensemble HCCR-GoogLeNet models achieve new state of the art recognition accuracy of 96.35% and 96.74%, respectively, outperforming previous best result with significant gap.**

## I. INTRODUCTION

The handwritten Chinese character recognition (HCCR) problem has been extensively studied for more than 40 years [1]-[7]. Nevertheless, HCCR is still a challenge unsolved problem owing to its large scale vocabulary (say, as many as 6763 classes in GB2312-80 standard, as large as 27533 in GB18010-2000 standard, or as huge as 70244 in GB18010-2005 standard), great diversity in handwriting styles (imaging there are more than 1 billion Chinese people), too many similar and confusable Chinese characters, and so on. It seems that traditional offline HCCR approaches, such as the modified quadratic discriminant function (MQDF) based methods, meet a bottleneck since no significant progress has been observed by us in recent years. The best traditional methods, such as MQDF or DLQDF [1][4][5], achieved a fairly good performance with less than 93% accuracy on a challenge offline HCCR database, CASIA-HWDB [1], leaving great margin with human performance.

With the blooming of deep learning in recent years [12], convolutional neural networks (CNN) brings about new break-through technology for HCCR with great success [6][11][12], narrowing the margin between these methods and human performance. The CNN, which was original developed in 1990s by LeCun [8][9], has been studied extensively in recent years. CNN has been extended using deeper architectures (c.f., [13], better training technologies such as such as Dropout [15] and better nonlinear activation function such as ReLU [12], to solve numerous computer vision challenges and pattern recognition problems with great significant success. Among them, the multi-column deep neural network (MCDNN) method proposed by Cireşan et al. [11][16], may be the first that reported the successful application of CNN for large vocabulary HCCR. However, the MCDNN is, in fact, a simple average voting ensemble model, composed of several standard CNNs. It is interesting to observe that deep convolutional neural network models continued their success by winning both online and offline handwritten Chinese character recognition competitions at the ICDAR 2013 [5].The team from Fujitsu won first place in the offline HCCR competition, with an accuracy rate of 94.77 %. In 2014, Wu et al. further improved the performance of offline HCCR to 96.06%, based on the voting of four alternately trained relaxation convolutional neural networks (ATR-CNN) [6].

Although significant progress has been achieved by using CNN-based models for offline HCCR, most existing models simply process handwritten Chinese characters as image patterns. Based on this, the CNN is directly applied as an end-to-end black box for HCCR, without using any vital domain-specific information such as feature extraction, which may be useful but cannot be learned by neural networks. Moreover, some previous models were neither deep enough nor slim enough. For example, the MCDNN is 10 layers deep and the ART-CNN is only 9 layers deep, when counting layers of both convolutional and pooling. Furthermore, the winner model of the Fujitsu team in the ICDAR offline HCCR competition required a dictionary storage size of 2.46 GB [6]; as a result, this model is less practically useful for mobile applications.

In this paper, based on a recently developed deep CNN model by Google, namely GoogLeNet [13], we present a new deep CNN based method for HCCR. The proposed CNN model is a streamlined version of GoogLeNet, where we use less *Inception* module than the original one. Moreover, we employ three types of directional feature maps, namely the Gabor, gradient and HoG feature maps, to enhance the performance of GoogLeNet.

The remainder of this paper is organized as followings. Section II gives a brief introduction of CNN. Section III presents two CNN models we designed for HCCR, one shallow model named HCCR-AlexNet, the other HCCR-GoogLeNet. The domain feature extraction methods are introduced in section IV. Experimental results are given in section V, and conclusions are drawn in section VI.

## II. A BRIEF INTRODUCTION OF CNN

CNNs [8][9] is a hierarchical neural networks that extract local features by convolving input with a group of kernel filters. The obtained convolutional feature maps are then sub-sampled (denoted as pooling) and filtered out to next layer. In the following, we will brief introduce CNN algorithm.

Given $x_i^l \in \mathbb{R}^{M_l \times M_l}$ represents the $i^{th}$ map in the $l^{th}$ layer, $j^{th}$ kernel filter in the $l^{th}$ layer connected to the $i^{th}$ map in the $(l-1)^{th}$ layer denoted $k_{ij}^l \in \mathbb{R}^{K_l \times K_l}$ and index maps set $M_j = \{ i \mid i^{th}$ in the $(l-1)^{th}$ layer map connected to $j^{th}$ map in the $l^{th}$ layer $\}$. So the convolution operation can be given by equation (1).

$$x_j^l = f(\sum_{i \in M_j} x_i^{l-1} * k_{ij}^l + b_j^l) \quad (1)$$

where $f(.)$ is ReLU non-linearity activation function $f(z) = \max(0, z)$, $b_j^l$ is bias. And pooling equation can be described in equation (2).

$$x_j^l = down(x_i^{l-1}) \quad (2)$$

where $down(.)$ is sum-sampling function to computer the max value of each $n \times n$ region in $x_i^{l-1}$ map.

Softmax regression is applied as an effective method for multi-class classification problem. Suppose we have $T$ categories and the training data for the each category are denoted as $(x_i, y_i)$, where $i = \{1, ..., N\}$, with $x_i \in \mathbb{R}^d$ and $y_i \in \mathbb{R}$ being the feature vector and label apart. CNN aims to minimize the following cross-entropy loss function:

$$J(\theta) = -\frac{1}{N}\left[\sum_{i=1}^{N}\sum_{t=1}^{T} 1\{y_i = \mathrm{t}\} \log \frac{e^{\theta_t^T x_i}}{\sum_{l=1}^{T} e^{\theta_l^T x_i}}\right] \quad (3)$$

where $\theta$ is model parameters, $\sum_{l=1}^{T} e^{\theta_l^T x_i}$ is a factor of normalization, $1(.)$ is an indicating function.

The loss function of $J(\theta)$ can be minimized by using stochastic gradient descent (SGD) algorithm, during the training process of CNN.

## III. DESIGN OF TWO CNNS FOR HCCR

Inspired by the outstanding work of Krizhevsky et al. [12] and Szegedy et al. [13], which won the first place at the ILSVRC-2012 contest and at the ILSVRC-2014 contest respectively, we designed two CNN architectures, named HCCR-AlexNet and HCCR-GoogLeNet for offline HCCR. The HCCR-AlexNet takes the same architecture as [12], which consists of eight weighted layers; the first five layers include three groups of convolutional layers and max-pooling layers as well as two single convolutional layers; the remaining three layers are fully connected layers, and the detail of HCCR-AlexNet is depicted in Figure 1.

Another CNN model we designed for HCCR follows the idea of GoogLeNet [13], which was the winner of ILSVRC-2014 [17]. One significant characteristic of GoogLeNet is that it is designed very deep, while the network is 22 layers deep when counting only layers with parameters (or 27 layers if also count pooling layers). Another characteristic of GoogLeNet is that a new local *Inception* module was introduced to CNN. The basic idea of *Inception* module is to find the optimal local construction and to repeat it spatially. One of the main beneficial aspects of this architecture is that it allows for increasing the number of units at each stage significantly without an uncontrolled blow-up in computational complexity. So that the CNN can be designed not only very deeply but also be efficiently trainable. Out model is shown in Figure 2, which is named as HCCR-GoogLeNet. It is 14 layers deep when counting only layers with parameters (or 19 layer deep if also count pooling layers and input layer and softmax output), consisted of 4 *inception* modules. Each *inception* modules is made up of $1\times1$ convolutions, $3\times3$ convolutions, $5\times5$ convolutions, and $3\times3$ max pooling. Moreover, $1\times1$ convolutions are applied to computer reductions for involving with less number of parameters as well as rectified activation before the expensive $3\times3$ and $5\times5$ convolutions. Thanks to the *inception* module, we can extract local feature representation using flexible convolutional kernel filter sizes with layer-by-layer structure, which was proved to be robust and effective for the large scale high-resolution images. Furthermore, owing to the padding strategy and precise designs, after the *inception* module operation, we can obtain a number of feature maps of the same sizes, though by means of different scale convolutions as well as pooling; and the feature maps are concatenated together by a concat-layer following by each inception module.

## IV. EMBEDDING OF DIRECTIONAL FEATURE MAPS TO HCCR-GOOGLENET

Feature extraction is an important step of traditional technique for HCCR, admitting that CNN is applied as an end-to-end neural network, which integrates feature extraction and classification are integrated together during the training processing. However, CNN is considered as a black box of learning for HCCR, neglecting some valid domain-specific information that cannot be learned by CNN. In this paper, we extracted directional feature maps as prior knowledge and added obtained feature maps into the input layer as well as the original image to enhance the performance of HCCR-GoogLeNet.

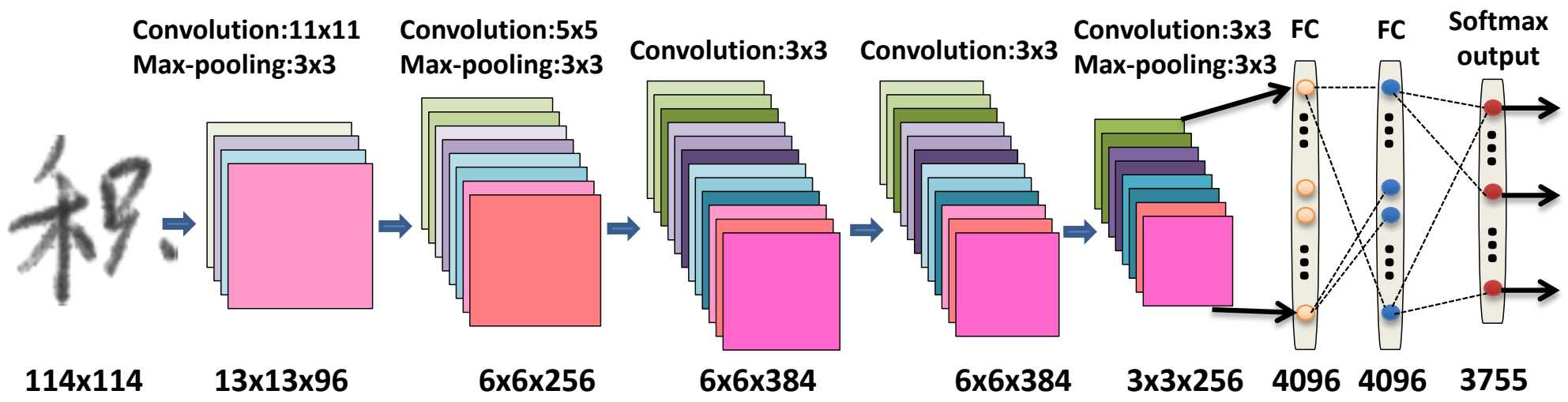

Figure 1. The architecture of HCCR-AlexNet for offline HCCR

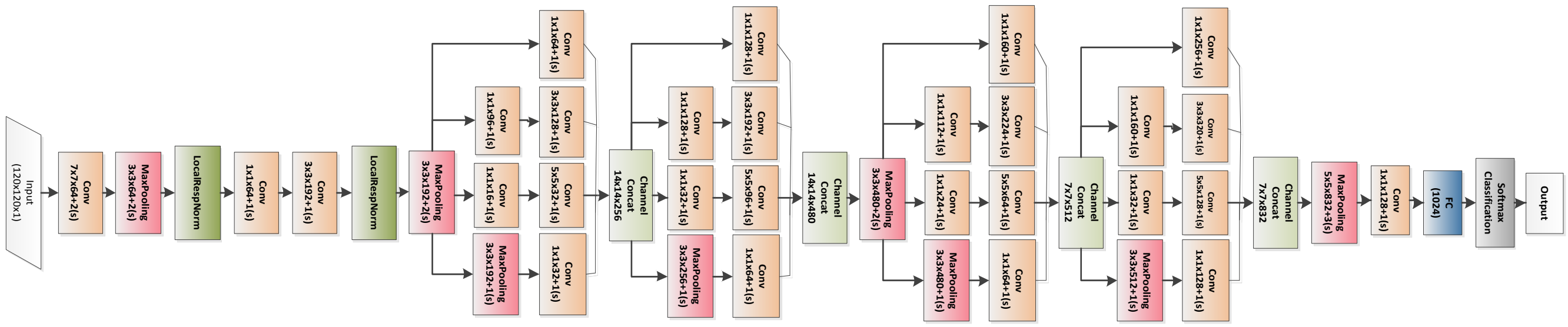

Figure 2. The architecture of our HCCR-GoogLeNet for offline HCCR

Gabor transformation is widely used in image process, and has achieved very promising results for offline handwritten Chinese character recognition [18, 21]. The multi-orientation Gabor transformation is given by:

$$F(x,y;\kappa,\vartheta_k) = I(x,y) * G(x,y;\kappa,\vartheta_k) \quad (4)$$

where $I(x,y)$ stands for input image, $G(x,y;\kappa,\vartheta_k)$ denotes Gabor filter. The detail of Gabor filter is:

$$G_1(x,y) = \frac{\kappa^2}{\sigma^2}\exp[-\frac{\kappa^2(x^2+y^2)}{2\sigma^2}] \quad (5)$$

$$\Re = \kappa x \cos\vartheta_k + \kappa y \sin\vartheta_k \quad (6)$$

$$G(x,y;\kappa,\vartheta_k) = \begin{array}{l} G_1(x,y)[\cos(\Re) - \exp(-\frac{\sigma^2}{2})] \\ +iG_1(x,y)\sin(\Re) \end{array} \quad (7)$$

where $\sigma = \pi$ $\kappa = \frac{2\pi}{\iota}$, $\vartheta_k = \frac{\pi k}{M}$ with $k = 0,1,2,...,M-1$. The parameters $\iota$ and $\vartheta_k$ are wavelength and orientation apart.

In our paper, we select $M = 8$, that is eight orientations corresponding to 0º, 22.5º, 45º, 67.5º, 90º, 112.5º, 135º, 157.5º; and we select the wavelength $\iota = 4\sqrt{2}$ after several empirical analysis. After Gabor feature extraction, we can obtain 8 different orientations within signal wavelength Gabor feature maps with the same size of each offline Chinese character image. The 8 Gabor feature maps are added into the input layer as well as the original image to construct an $N \times N \times 9$ CNN-input-layer array, where $N \times N$ stands for the size of input character image.

Moreover, gradient feature is proved to be effective in HCCR [20], we also apply gradient feature maps in the same way. The Sobel operator is used to computer the gradient value of each pixel in the input image on the x-axis and y-axis respectively, then we decompose each gradient value vector to eight direction. That is, after gradient feature extraction, we obtain 8 gradient feature maps, which are put into the input layer with the primary image. In addition, HoG is proposed as a good feature in computer vision [19], so we also utilize HoG feature for HCCR. Figure 3 illustrate the Gabor, gradient, and HoG feature maps of a handwritten Chinese character “积” .

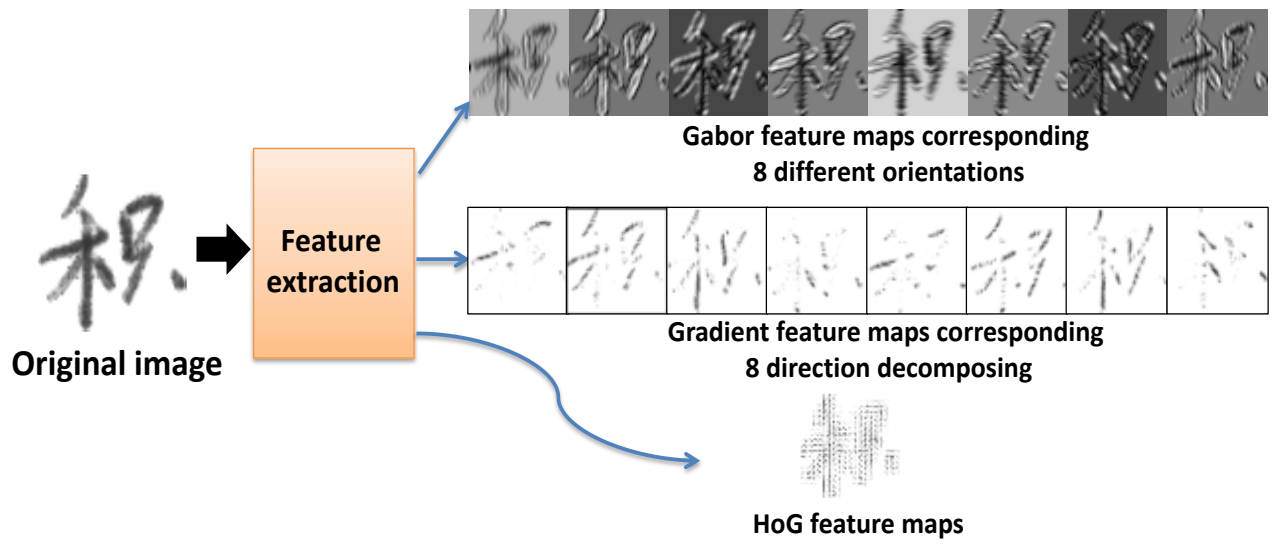

Figure 3. Illustratio of Gabor, gradient, and HoG feature maps of an given offline Chinese character “积”

## V. Experiments and Analysis

### A. Experimental Data

We used the offline CASIA-HWDB1.0 (DB1.0) and CASIA-HWDB1.1 (DB1.1) [1] databases for training, and the test dataset from the 2013 ICDAR Chinese handwriting recognition competition (noted as CompetitionDB) [5] for testing, which were all collected by the Institute of Automation of the Chinese Academy of Sciences. DB1.0 contains 3,740 Chinese characters in the GB2312-80 standard level-1 set, which were contributed by 420 writers; the DB1.1 contains 3,755 classes in GB1, contributed by 300 writers. The CompetitionDB contains 3,755 classes contributed by another 60 writers.

*B. Pre-processing and experimental settings*

We shuffle our training data firstly, after visual inspection of several characters rescaled to various sizes and suitable to our designed experimental CNN architecture, we decided to normalize offline characters image to a size of 108x108 for HCCR-AlexNet and 112×112 for HCCR-GoogLeNet. Before resizing, we reversed the grey values of the image in order to ensure fast computation. Then, we placed a scaled character at the center of the 114×114 pixels mask for HCCR-AlexNet and 120×120 pixels mask for HCCR-GoogLeNet so that the margin information is maintained. We conducted our experiments on an open CNN platform called Caffe [22] using a GTX TITAN BLACK GPU card.

*C. Comparison of HCCR-GoogLeNet against HCCR-AlexNet*

AlexNet [12] was the first success CNN model to have a good performance for large scale high-resolution image, and GoogLeNet [13] with *inception* modules and deeper layer-by-layer convolutions structure have sharply improved the recognition rate in ILSVRC-2014 contest. We evaluated the performance comparison between our designed HCCR-AlexNet and HCCR-GoogLeNet. For the testing 3755 classes of CompetitionDB, the recognition rate with AlexNet we obtained was 95.49%, whereas the recognition rate with HCCR-GoogLeNet was 96.26%. It can be seen that HCCR-GoogLeNet improved the recognition performance significantly, which demonstrates that deeper layer-by-layer architecture help to extract the more nature and abstract representation of offline Chinese characters and improve the accuracy. Figure 4 shows a comparison of these two models in terms of the testing recognition accuracy.

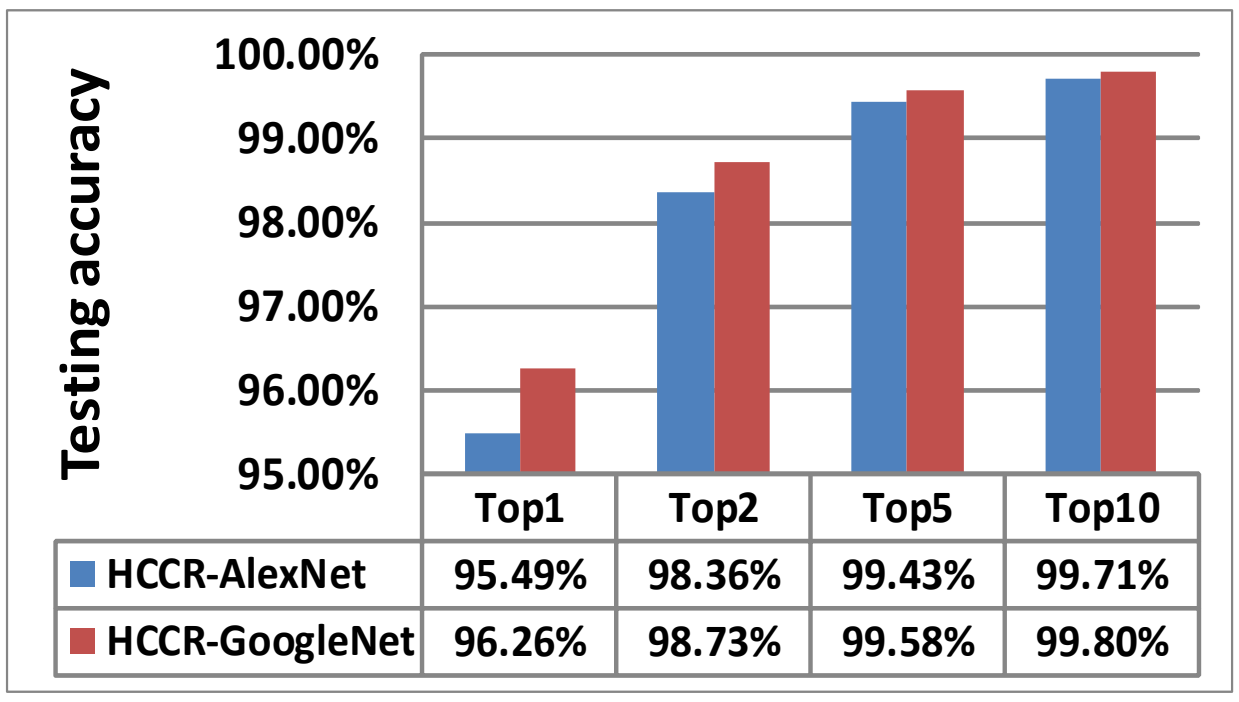

Figure 4. Comparison between HCCR-AlexNet and HCCR-GoogLeNet

*D. Results of HCCR-GoogLeNet plus Gabor for HCCR*

In these experiments, we compare performance of Gabor, gradient, and HoG feature maps as prior knowledge embedding into the input layer of HCCR-GoogLeNet, as well as their ensemble performance. The corresponding models are denoted as Gabor+HCCR-GoogLeNet, Gradient+HCCR-GoogLeNet, HoG+HCCR-GoogLeNet respectively. The ensemble method we used is average of several inputting model. Two ensemble models were designed by us, one is just average the above four HCCR-GoogLeNet with different type of feature maps, the other is to ensemble of 10 HCCR-GoogLeNet using six additional models including the HCCR-AlexNet and five HCCR-GoogLeNet with different usage of feature maps and training data. The results are listed in Table I. It can be seen that the recognition rate of Gabor+HCCR-GoogLeNet was 96.35%, which is superior to Gradient+HCCR-GoogLeNet, HoG+HCCR-GoogLeNet and HCCR-GoogLeNet. This indicates clearly that the effectives of using Gabor feature maps can further improve the performance of GoogLeNet for HCCR. We also conducted an experiment using only 8-directional Gabor feature maps into the input layer without the origin input image bitmap, and the recognition rate in this case was 95.08%, demonstrating the valid representation for character just via Gabor feature maps. Furthermore, we combined four and ten pre-trained HCCR-GoogLeNet models, and we got a higher accuracy of 96.64% and 96.74% respectively.

TABLE I. RECOGNITION ACCURACIES OF HCCR-GOOGLENET-CNNS USING DIFFERENT DIRECTIONAL FEATURES

| Methods | Top1 (%) | Top2 (%) | Top5 (%) | Top10 (%) |
|---|---|---|---|---|
| HCCR-GoogLeNet | 96.26 | 98.73 | 99.58 | 99.80 |
| HoG+HCCR-GoogLeNet | 96.25 | 98.73 | 99.56 | 99.80 |
| **Gabor+HCCR-GoogLeNet** | **96.35** | **98.75** | **99.60** | **99.80** |
| Gradient+HCCR-GoogLeNet | 96.28 | 98.72 | 99.56 | 99.80 |
| **HCCR-Ensemble-GoogLeNet (average of 4 models)** | **96.64** | **98.90** | **99.64** | **99.83** |
| **HCCR-Ensemble-GoogLeNet (average of 10 models)** | **96.74** | **98.93** | **99.65** | **99.83** |

*E. Comparison of different methods on the CompetitionDB*

To show the outstanding performance the proposed HCCR-GoogLeNet CNN models, we compare the performances of different methods using the ICDAR 2013 offline HCCR competition datasets. The results are given in Table II. It can be seen that our single and ensemble HCCR-GoogLeNet model (average of 4 or 10 single models) outperform all others. Compared with the best results of the traditional MQDF-based approach (DLQDF) [5], our method (HCCR- GoogLeNet) achieved a significant improvement with a relative 55.22% error rate reduction. Comparing with the winner of ICDAR 2013 offline Chinese character recognition competition (CNN-Fujitsu), our method achieved a significant improvement with a relative 37.67% error rate reduction. Compared with the best state-of-the-art CNN-based result (ATR-CNN Voting) [6], we also obtained a relative 17.26% error rate reduction. It is worth noting that the ATR-CNN contains only 10 layers, but involves approximate 12.91 million parameters (or 51.64MB storage), in comparison, our HCCR-GoogLeNet has 19 layers deep but only involves with 7.26 million parameters (or 27.68MB) to achieve a better recognition performance. This confirms us that the much deeper yet slim architecture of HCCR-GoogLeNet is a very promising model for higher performance recognition of handwritten Chinese character.

TABLE II. RECOGNITION ACCURACIES OF OUR PROPOSED METHOD AGAINST PREVIOUS METHODS

| System | Top1 (%) | Top5 (%) | Top10 (%) | Dic. size |
|---|---|---|---|---|
| **HCCR-Gabor-GoogLeNet** | **96.35** | **99.60** | **99.80** | **27.77MB** |
| **HCCR-Ensemble-GoogLeNet** (average of 4 models) | **96.64** | **99.64** | **99.83** | **110.91MB** |
| CNN-Fujitsu[5] | 94.77 | n/a | 99.59 | 2460MB |
| MCDNN-INSIA [5] | 94.42 | n/a | 99.54 | 349MB |
| MQDF-HIT [5] | 92.61 | n/a | 98.99 | 120MB |
| MQDF-THU [5] | 92.56 | n/a | 99.13 | 198MB |
| DLQDF [5] | 92.72 | n/a | n/a | n/a |
| ART-CNN [6] | 95.04 | n/a | n/a | 51.64M* |
| R-CNN Voting [6] | 95.55 | n/a | n/a | 51.64M* |
| ATR-CNN Voting [6] | 96.06 | n/a | n/a | 206.56MB* |

*This value is estimated by us according to the CNN parameters given by [6]

## VI. CONCLUSION

In this paper, we present a new deep leaning model, HCCR-GoogLeNet, for the recognition of handwritten Chinese character. HCCR-GoogLeNet uses four *Inception* modules to construct an efficient deep network, taking the advantage of finding optimal local construction and allowing to repeat it spatially. The HCCR-GoogLeNet is designed very deeply yet slim, with total 19 layers (counting for all convolutional layers, pooling layers, fully connect layers and softmax output layer). We employ directional feature extraction domain knowledge, such as the Gabor feature, HoG feature and gradient feature, to enhance the performance of HCCR-GoogLeNet. Experiments on the ICDAR 2013 offline HCCR competition dataset show that our best single HCCR-GoogLeNet is superior to all previous best single and ensemble CNN models in terms of both accuracy and storage performance. The best testing error rate we achieved is 3.26%, which is a new state-of-the-art record to our best knowledge.

## ACKNOWLEDGEMENT

This research is supported in part by NSFC (Grant No.: 61472144), National science and technology support plan (Grant No.:2013BAH65F01-2013BAH65F04), GDSTP (Grant No.: 2012A010701001, 2012B091100396). We gratefully appreciate the support of NVIDIA Corporation with the donation of GPUs used for this research.